\documentclass{article}

\PassOptionsToPackage{numbers, compress}{natbib}

\usepackage[final, main]{neurips_2025}

\usepackage[utf8]{inputenc}
\usepackage[T1]{fontenc}
\usepackage{hyperref}
\usepackage{url}
\usepackage{booktabs}
\usepackage{amsfonts}
\usepackage{nicefrac}
\usepackage{microtype}
\usepackage{xcolor}
\usepackage{listings}

\title{HiFi-RAG: Hierarchical Content Filtering and Two-Pass Generation for Open-Domain RAG}

\author{%
  Cattalyya Nuengsigkapian \\
  Google \\
  \texttt{cattalyya@google.com}
}

\begin{document}

\maketitle

\begin{abstract}
  Retrieval-Augmented Generation (RAG) in open-domain settings faces significant challenges regarding irrelevant information in retrieved documents and the alignment of generated answers with user intent. We present \textbf{HiFi-RAG} (Hierarchical Filtering RAG), the winning closed-source system in the Text-to-Text static evaluation of the MMU-RAGent NeurIPS 2025 Competition. Our approach moves beyond standard embedding-based retrieval via a multi-stage pipeline. We leverage the speed and cost-efficiency of Gemini 2.5 Flash (4-6$\times$ cheaper than Pro) for query formulation, hierarchical content filtering, and citation attribution, while reserving the reasoning capabilities of Gemini 2.5 Pro for final answer generation. On the MMU-RAGent validation set, our system outperformed the baseline, improving ROUGE-L to 0.274 (+19.6\%) and DeBERTaScore to 0.677 (+6.2\%). On \textbf{Test2025}, our custom dataset evaluating questions that require post-cutoff knowledge (post January 2025), HiFi-RAG outperforms the parametric baseline by 57.4\% in ROUGE-L and 14.9\% in DeBERTaScore.
\end{abstract}

\section{Introduction}
The MMU-RAGent (Massive Multi-Modal User-Centric Retrieval Augmented Generation Benchmark) competition challenges participants to build systems that retrieve from web-scale corpora to answer diverse user queries. A common failure mode in such systems is the retrieval of irrelevant context which causes hallucination ("garbage-in, garbage-out").

Our solution, HiFi-RAG, prioritizes \textit{precision} in the context window. We abandon standard vector-similarity search in favor of a hierarchical filtering approach. Similar to multi-stage ML model cascades that utilize low-power signals to gate high-power processing \cite{cascade}, we employ Gemini 2.5 Flash \cite{gemini} as a lightweight gatekeeper to semantically filter hierarchically parsed web content before forwarding it to the more computationally intensive Gemini 2.5 Pro. This ensures the deep reasoning model receives only the most salient information, significantly reducing computational load.

\section{Methodology}

Our pipeline consists of five distinct stages: Query Planning, Retrieval, Hierarchical Filtering, Two-Pass Generation, and Citation Verification.

\subsection{Query Formulation}
User queries are often too verbose or conversational for effective retrieval (e.g., asking for an "ELI5" explanation). We utilize Gemini 2.5 Flash to analyze the user intent and propose optimized search queries. We explicitly instruct the model to \textit{"Create an effective and concise Google search query"} (see full prompt in Appendix \ref{app:queryprompt}). As shown in Table \ref{tab:query_examples}, using samples from the MMU-RAGent validation set, this step extracts core intent and distinct search terms, improving recall for complex constraints.

\begin{table}[h]
  \caption{Examples of Query Formulation (Inputs from MMU-RAGent Validation Set)}
  \label{tab:query_examples}
  \centering
  \small
  \begin{tabular}{p{0.48\linewidth}p{0.45\linewidth}}
    \toprule
    \textbf{User Input (Raw)} & \textbf{Generated Search Queries} \\
    \midrule
    You are a media technology professor with 20 years of experience. Explain to me like I am five, how a camera works. & \texttt{['how a camera works explained for 5 year old', 'ELI5 how a camera works']} \\
    \midrule
    how do i stack up 10 four-feet chairs vertically so that it remains stable? & \texttt{['how to stack chairs safely', 'stable chair stacking techniques']} \\
    \midrule
    Tell me the difference between "the office" and "modern family" about the actors looking at the camera & \texttt{["The Office" "Modern Family" looking at camera difference', '"The Office" "Modern Family" fourth wall comparison']} \\
    \bottomrule
  \end{tabular}
\end{table}

\subsection{Retrieval and URL Filtering}
Upon receiving initial Google Search API results, we employ a pre-fetch filtering step. Instead of scraping every result, Gemini Flash analyzes the URL, title, and preview content from Search API to select only the most relevant sources (see Appendix \ref{app:urlfilteringprompt}). This process reduces the URL count by 33.5\% (averaging across 100 queries). By proactively discarding irrelevant domains (e.g., gaming vs. aerospace), outdated information, mismatched contexts, missing key constraints, and speculative discussions before expensive scraping, we improve both latency and context quality.

\subsection{Hierarchical Content Parsing \& Filtering}

\textbf{Hierarchical Content Parsing:} We utilize the Scrapingdog API (web) and Reddit API (forums) to handle structural complexity.
\begin{itemize}
\item \textbf{Hierarchical Parsing:} Rather than treating content as flat text, we parse HTML into hierarchical sections (chunks). Text blocks are explicitly grouped under their parent headers (e.g., \texttt{<h1>}--\texttt{<h4>}) as a markdown plain-text.
\item \textbf{Reddit Tree Reconstruction:} Preserves discussion flow by retrieving top-$k$ comments with top-$m$ nested replies across two layers $(k=5, m_1=3, m_2=2)$.
\end{itemize}

\textbf{Section filtering and ranking (LLM-as-a-Reranker):} Instead of embedding-based filtering, we deploy Gemini 2.5 Flash to evaluate each parsed section against the user query, using only its title and a small snippet (the first 200 characters of content) to make the evaluation as lightweight as possible. It ranks sections by relevance using a custom prompt (see Appendix \ref{app:chunkfilteringprompt}) and discards noise. This removes ~60.5\% of chunks (averaged across 100 queries), resulting in a context window dense with high-quality signals.

\subsection{Two-Pass Generation (Gemini Pro)}
We utilize \textbf{Gemini 2.5 Pro} in a two-turn conversation to separate factuality from style (see Appendix \ref{app:generationprompt}):
\begin{enumerate}
    \item \textbf{Turn 1 (Drafting):} The model generates a comprehensive answer from filtered content, where each site contains title, url, preview, and filtered sections sorted by LLM relevance.
    \item \textbf{Turn 2 (Refinement):} The model is prompted to revise its answer to match the style and length of three hand-picked question-answering examples from the validation set, distinct from the first 100 pairs used for evaluation (e.g., step-by-step guides for "how-to" questions).
\end{enumerate}

\subsection{Post-Hoc Citation Verification}
To ensure attribution accuracy, we employ a dedicated verification step using Gemini 2.5 Flash. We decouple citation from generation to prevent performance degradation in both answers and citations caused by long context windows. This allows the verification step to focus exclusively on source attribution and prioritizing high-quality sources when duplicates exist (see Appendix \ref{app:citationprompt}) to provide source context indices that \textit{directly support} the claims.

\section{Experiments and Results}

\subsection{Experimental Setup}
The competition provided an official validation set (with ground truth) but no training set, alongside a final blind test set without answers. Consequently, we adopted a few-shot approach and utilized the MMU-RAGent Text-to-Text validation set as our primary development benchmark. Due to cost and time constraints during the ablation phase, we evaluated on the first 100 (out of 300) samples. Evaluation was performed using ROUGE-L \cite{rouge} and DeBERTaScore \cite{deberta}. For DeBERTaScore, we utilized the `microsoft/deberta-xlarge-mnli` model \cite{debertahf} to ensure robust semantic matching.

Additionally, we evaluated on a custom \textbf{Test2025} dataset (100 samples). Test2025 was synthesized to evaluate retrieval performance on events occurring after February 2025, strictly enforcing a scenario where the model must rely on retrieved context rather than parametric memory.

\subsection{Main Results (Standard Validation)}
We performed an ablation study to quantify the impact of each pipeline component. Table \ref{tab:results} summarizes the performance progression.

For the \textbf{Baseline Q} (Raw Query) configuration, we queried Gemini Pro with the user's query directly, but also provide word limit to ensure fair comparison with the ground truth:  \textit{"Please limit your answer to under 200 words. [USER\_QUERY]"}.

\begin{table}[h]
  \caption{Ablation on MMU-RAGent Standard Validation Set}
  \label{tab:results}
  \centering
  \small
  \begin{tabular}{lcc}
    \toprule
    System Configuration & ROUGE-L (F1) & DeBERTaScore (F1) \\
    \midrule
    Baseline Q (Raw Query + Length Constraint) & 0.2291 & 0.6375 \\
    Baseline Prompt (No Search) & 0.2591 & 0.6667 \\
    RAG (Search enabled) & 0.2664 & 0.6677 \\
    RAG w/ Filters (URL + Chunk) & 0.2695 & 0.6712 \\
    \textbf{Final (RAG w/ Filters + Rephrase + 2-Turn)} & \textbf{0.2739} & \textbf{0.6772} \\
    \bottomrule
  \end{tabular}
\end{table}

The final configuration provided a 19.6\% and 6.2\% improvement over the baseline on ROUGE-L and DeBERTaScore respectively. Notably, prompt engineering alone (Baseline Prompt) provided a significant 13\% improvement on ROUGE-L, highlighting the importance of instructional clarity. The addition of Search (RAG) and active Filtering (Filters) added another 4\%, with the full two-turn refinement system achieving the highest scores across all metrics.

\subsection{Retrieval Evaluation on Future Events (Test2025)}
To strictly evaluate the system's retrieval capabilities rather than parametric memory, we prompted Gemini 3.0 (Thinking mode with web search) to create the Test2025 dataset, containing question-answering pairs for knowledge that became available after February 2025. Note that as of November 2025, Gemini 2.5 Pro and Gemini 2.5 Flash, used in our system (as well as Gemini 3.0), all currently share a knowledge cutoff of January 2025.

\begin{table}[h]
  \caption{Performance on Test2025 (Future Events > Feb 2025)}
  \label{tab:val2025}
  \centering
  \small
  \begin{tabular}{lcc}
    \toprule
    System Configuration & ROUGE-L (F1) & DeBERTaScore (F1) \\
    \midrule
    Baseline Q (Raw Query) & 0.2022 & 0.6173 \\
    Baseline Prompt & 0.2766 & 0.6574 \\
    RAG (Search enabled) & 0.2915 & 0.6776 \\
    RAG w/ URL Filter Only & 0.2966 & 0.6829 \\
    RAG w/ Filters (URL + Chunk) & 0.3031 & 0.6840 \\
    RAG w/ Filters + Rephrase & 0.2898 & 0.6832 \\
    \textbf{Final (RAG w/ Filters + Rephrase + 2-Turn)} & \textbf{0.3182} & \textbf{0.7092} \\
    \bottomrule
  \end{tabular}
\end{table}

On this dataset, Baseline Q performance drops significantly (0.2022). Consequently, the performance gap between the configuration with web corpora (RAG) and the one without (Baseline Q) widens dramatically from \textbf{16.3\%} on the MMU-RAGent validation set to \textbf{44.16\%} on Test2025, confirming the knowledge cutoff limitation.

Our RAG configurations restore performance, effectively bridging the gap to the present day and increasing ROUGE-L and DeBERTaScore by \textbf{57.4\%} and \textbf{14.9\%}, respectively, compared to the baseline. For the final test, we replaced the three one-shot examples (originally drawn from the validation set) with three new pairs generated by Gemini 3.0 to ensure style alignment with the test set distribution. 

Furthermore, we observed that query rephrasing without URL filtering (RAG w/ Filters + Rephrase) resulted in lower scores, likely due to the removal of context from verbose queries which can introduce ambiguity. Because URL filtering mitigates these errors, query refinement proves most effective when implemented in conjunction with filtering.

\subsection{Negative Results}
We explored several alternative strategies that were discarded due to poor performance or high cost.
\begin{itemize}
    \item \textbf{Embeddings vs. LLM Filtering:} We attempted to filter content using sentence embeddings (Voyage AI \cite{voyage}). This performed worse than LLM-based filtering, likely because embeddings struggled to distinguish between topically related but factually irrelevant noise.
    \item \textbf{Agentic Workflows:} We implemented a full Gemini Agent with search tools. This approach was $10\times$ more expensive and significantly slower, often timing out without improving ROUGE scores compared to our deterministic pipeline.
    \item \textbf{DSPy Optimization:} We used DSPy \cite{dspy} with the GEPA evolutionary optimizer \cite{gepa} to tune prompts. The optimizer tended to overfit to the validation set, producing brittle prompts that failed to generalize.
    \item \textbf{LLM-as-a-Judge Refinement:} A "Checker" module critiqued answers to catch hallucinations. While qualitatively better, it degraded automated metrics (ROUGE/DeBERTaScore), likely by altering the phrasing too aggressively away from reference text.
\end{itemize}

\section{Conclusion}
HiFi-RAG demonstrates that a structured, multi-stage pipeline using  LLMs for filtering and generation outperforms traditional agentic approaches for open-domain RAG. By leveraging Gemini 2.5 Flash for high-throughput filtering and Gemini 2.5 Pro for reasoning, we achieved a balance of cost, latency, and accuracy suitable for the MMU-RAGent benchmark.

\begin{ack}
We thank the MMU-RAGent organizers for providing the benchmark and putting this competition together.
\end{ack}


\newpage
\appendix

\section{Prompts}
\label{app:prompts}

\subsection{Query Refinement Prompt}
\label{app:queryprompt}
We prompt Gemini 2.5 Flash with the following instruction to transform user queries into search-engine-friendly keywords:
\begin{verbatim}
Create an effective and concise Google search query for this question: 
[USER_QUESTION]
Return a json list of strings for the best 1-2 search queries.
\end{verbatim}

\subsection{URL Filtering Prompt}
\label{app:urlfilteringprompt}
Before scraping, we filter the search results using Gemini 2.5 Flash with the following prompt to identify high-value targets:
\begin{verbatim}
What URLs from the list below would be helpful to read further to answer 
"[USER_QUESTION]"?
Please return a JSON list of URL strings. Here are the urls with their 
preview content:

[SEARCH_RESULT]
\end{verbatim}

\subsection{Chunk Filtering \& Ranking Prompt}
\label{app:chunkfilteringprompt}
The following prompt is used to select relevant sections based on their title and a preview of their content:
\begin{verbatim}
Given a webpage preview and its section titles and an opening snippet, 
help determine what sections are helpful for us to read further to 
help answer [USER_QUESTION] without having to search/research further.
Return a JSON list of the useful section indices, sorted by most usefulness first.

Example output: [3, 2, 6, 7]

==================
Webpage overview: [WEB_PREVIEW_CONTENT]
------------------
Section previews in the page: [SECTION_PREVIEWS]
------------------
Useful chunks:
\end{verbatim}

\subsection{Two-Turn Generation Prompts}
\label{app:generationprompt}
\textbf{Turn 1 (Drafting):}
\begin{verbatim}
You are a helpful and knowledgeable assistant.
Answer the user question in a plain text in one paragraph (1-4 sentences).
Include only the answer without any introductory phrases, conversational filler, 
or preamble.

User question: [USER_QUESTION]
-----------
Here're extra information from Web Search, you might find helpful:
[WEB_CONTENT]
-----------
[USER_QUESTION]
\end{verbatim}

\textbf{Turn 2 (Refinement):}
\begin{verbatim}
Revise your answer to have a style and length similar to the "answer" 
in the following examples:
[VAL_EXAMPLES]
\end{verbatim}

\subsection{Citation Verification Prompt}
\label{app:citationprompt}
We use a separate LLM call to extract citations, ensuring they strictly support the generated answer:
\begin{verbatim}
Read the ANSWER and identify which SOURCES (by [number]) directly support 
the information it contains (for citations purpose).
Only list indices of the sources that directly support the answer. 
If no sources match, return [].
If multiple sources support the same fact, prioritize the source that is 
the most specific and direct match.

Your output MUST be a single, valid JSON array of source indices.
Example Output: [1, 4, 6]

-----------
ANSWER: [AI_ANSWER]
-----------
SOURCES:
[WEB_CONTENT]
\end{verbatim}

\subsection{Test2025 Dataset Generation Prompt}
\label{app:prompts2025}
We prompted Gemini 3.0 Thinking (via the web interface with active Google Search access) to generate 100 question-answer pairs of knowledge that became public after their knowledge cutoff:

\begin{verbatim}
I want question and answer pairs in ".jsonl" format, similar to the 
style/length below, that focus on questions that test RAG-based LLM systems
for knowledge that became public after February 2025. 
Give me 100 question-answering pairs

----- Example .jsonl ----
[FIRST_10_MMURAG_VALSET]
\end{verbatim}

\section{Qualitative Analysis and Examples}
\label{app:examples}

\subsection{Query Formulation Outputs}
We provide additional examples of how Gemini 2.5 Flash rephrases user input (from the MMU-RAGent Validation Set) into search-optimized queries. This step is critical for handling conversational nuances ("I live on my own") or specific formatting requests ("step by step").

\begin{itemize}
    \item \textbf{Input:} "I live on my own. I need to tie my shoelaces. How do I do that?" \\
    \textbf{Queries:} \texttt{['how to tie shoelaces', 'shoelace tying guide']}
    
    \item \textbf{Input:} "i have just got a new puppy. how should i train it to walk on a leash? give me step by step instructions" \\
    \textbf{Queries:} \texttt{['puppy leash training step by step', 'how to train puppy to walk on leash']}
    
    \item \textbf{Input:} "how to remove unwanted people from a photo in photoshop" \\
    \textbf{Queries:} \texttt{['photoshop remove person from photo', 'photoshop remove unwanted objects']}
    
    \item \textbf{Input:} "Tell me how to dump my own bought games on switch and run them in an emulator" \\
    \textbf{Queries:} \texttt{['dump nintendo switch games for emulation guide', 'how to dump switch games yuzu ryujinx']}
\end{itemize}

\subsection{Test2025 Analysis: The Knowledge Cutoff Gap}
Test2025 highlights errors caused by the model's January 2025 knowledge cutoff. These examples show how HiFi-RAG bridges this gap, alongside instances where new knowledge is unnecessary and the baseline remains competitive.

\paragraph{Case 1: RAG Success (New Events)} 
In these examples, the Baseline correctly identifies that it lacks the information or hallucinates a refusal, whereas HiFi-RAG retrieves the correct answer aligned with the ground truth.

\begin{itemize}
    \item \textbf{Query:} \textit{Who is the halftime show performer for Super Bowl LIX in 2025?}
    \begin{itemize}
        \item \textbf{Reference:} Kendrick Lamar headlined the Apple Music Super Bowl LIX Halftime Show on February 9, 2025, at the Caesars Superdome in New Orleans, Louisiana.
        \item \textbf{Baseline:} "The performer... has not yet been announced. Any names currently circulating are purely speculation." [Incorrect/Refusal]
        \item \textbf{HiFi-RAG:} "Kendrick Lamar is the headliner for the Super Bowl LIX halftime show in 2025, joined by special guest SZA..." [Correct]
    \end{itemize}
    
    \item \textbf{Query:} \textit{What are the new features of the iPhone 17 Air released in September 2025?}
    \begin{itemize}
        \item \textbf{Reference:} Apple released the iPhone 17 'Air' in September 2025... Key features include a significantly thinner aluminum chassis, a single 48MP rear camera, the A19 chip...
        \item \textbf{Baseline:} "Apple has not announced or released an iPhone 17 Air... Any details currently available would be based purely on unconfirmed rumors." [Incorrect/Refusal]
        \item \textbf{HiFi-RAG:} "The iPhone Air, released in September 2025, features the thinnest-ever iPhone design... and is powered by the A19 Pro chip..." [Correct]
    \end{itemize}
\end{itemize}

\paragraph{Case 2: RAG Correction (Dynamic Updates)}
Here, the Baseline relies on outdated "future plans" known before the cutoff, while HiFi-RAG retrieves the updated reality (e.g., a delay).

\begin{itemize}
    \item \textbf{Query:} \textit{Did the Artemis II mission launch in 2025?}
    \begin{itemize}
        \item \textbf{Reference:} No, the Artemis II mission did not launch in 2025. While originally targeted for late 2025, NASA delayed the mission to no earlier than April 2026...
        \item \textbf{Baseline:} "The Artemis II mission has not yet launched. NASA is currently targeting a launch date of no earlier than September 2025." [Outdated Info]
        \item \textbf{HiFi-RAG:} "No, the Artemis II mission did not launch in 2025... delayed to a target date of no later than April 2026." [Correct Update]
    \end{itemize}
\end{itemize}

\paragraph{Case 3: Baseline Competence (Long-term Planning)}
Events planned years in advance (e.g., World Expos, End-of-Life cycles) are often correctly answered by the Baseline, proving that "future" questions do not always require RAG if the schedule is static.

\begin{itemize}
    \item \textbf{Query:} \textit{Where is the 2025 World Expo held and what is its theme?}
    \begin{itemize}
        \item \textbf{Reference:} Expo 2025 is held in Osaka, Kansai, Japan... The official theme of the expo is 'Designing Future Society for Our Lives'.
        \item \textbf{Baseline:} "The 2025 World Expo will be held in Osaka, Kansai, Japan. Its theme is 'Designing Future Society for Our Lives'." [Correct]
        \item \textbf{HiFi-RAG:} "The 2025 World Expo will be held in Osaka, Japan... centered on the theme 'Designing Future Society for Our Lives'..." [Correct]
    \end{itemize}
    \item \textbf{Query:} \textit{When did Windows 10 officially reach its end of support?}
    \begin{itemize}
        \item \textbf{Reference:} Microsoft officially ended support for Windows 10 Home and Pro editions on October 14, 2025.
        \item \textbf{Baseline:} "Windows 10 is scheduled to officially reach its end of support on October 14, 2025." [Correct]
        \item \textbf{HiFi-RAG:} "Windows 10 officially reaches its end of support on October 14, 2025..." [Correct]
    \end{itemize}
\end{itemize}

\end{document}